\documentclass{article}
\usepackage{arxiv}
\usepackage[utf8]{inputenc} % allow utf-8 input
\usepackage[T1]{fontenc}    % use 8-bit T1 fonts
\usepackage{url}            % simple URL typesetting
\usepackage{booktabs}       % professional-quality tables
\usepackage{amsfonts}       % blackboard math symbols
\usepackage{nicefrac}       % compact symbols for 1/2, etc.
\usepackage{microtype}      % microtypography
\usepackage{lipsum}
\usepackage{lscape}
\usepackage{booktabs}
\usepackage{multirow}
\usepackage{graphicx}

\newcommand\blfootnote[1]{%
  \begingroup
  \renewcommand\thefootnote{}\footnote{#1}%
  \addtocounter{footnote}{-1}%
  \endgroup
}
\title{A Comparison of Pre-trained Vision-and-Language Models for Multimodal Representation Learning across Medical Images and Reports}
\author{
  Yikuan Li\\
  Feinberg School of Medicine\\
  Northwestern University\\
  Chicago, IL 60611\\
  \texttt{yikuanli2018@u.northwestern.edu}
   \And
 Hanyin Wang\\
  Feinberg School of Medicine\\
  Northwestern University\\
  Chicago, IL 60611\\
  \texttt{hanyinwang2022@u.northwestern.edu}
   \And
 Yuan Luo\thanks{Corresponding author}\\
  Feinberg School of Medicine\\
  Northwestern University\\
  Chicago, IL 60611\\
  \texttt{yuan.luo@northwestern.edu}
  }
\begin{document}
\maketitle
\begin{abstract}
Joint image-text embedding extracted from medical images and associated contextual reports is the bedrock for most biomedical vision-and-language (V+L) tasks, including medical visual question answering, clinical image-text retrieval, clinical report auto-generation. In this study, we adopt four pre-trained V+L models: LXMERT, VisualBERT, UNIER and PixelBERT to learn multimodal representation from MIMIC-CXR radiographs and associated reports. The extrinsic evaluation on OpenI dataset shows that in comparison to the pioneering CNN-RNN model, the joint embedding learned by pre-trained V+L models demonstrate performance improvement in the thoracic findings classification task. We conduct an ablation study to analyze the contribution of certain model components and validate the advantage of joint embedding over text-only embedding. We also visualize attention maps to illustrate the attention mechanism of V+L models.
\end{abstract}

% keywords can be removed
% \keywords{First keyword \and Second keyword \and More}

\section{Introduction}
There\blfootnote{Our implementation can be found at: https://github.com/YIKUAN8/Transformers-VQA} has been a long history of learning visual and semantic information to solve vision and language tasks. Various pioneering approaches \cite{yu2017multi,aneja2018convolutional,lee2018stacked,perez2018film,rohrbach2016grounding} have been proposed to boost the performance of visual question answering (VQA) \cite{antol2015vqa}, image captioning, image-text matching, visual reasoning \cite{karpathy2015deep} and textual grounding, respectively. These models learned image representations from powerful backbone convolutional neural networks (CNN), text representations from recurrent neural networks (RNN), and a fusion modality to achieve joint image-text embedding. For simplicity, these genres of models are often referred to as CNN+RNN models. Most of these models were designed for specific vision-an-language tasks and therefore had relatively poor generalizability. Recently, inspired by the success of large-scale pre-trained language models, such as BERT \cite{DBLP:journals/corr/abs-1810-04805}, XLNET \cite{DBLP:journals/corr/abs-1906-08237}, more researchers focus on generating image-text join embedding from pre-training Transformer-based \cite{DBLP:journals/corr/VaswaniSPUJGKP17} model on V+L datasets. The joint embedding is then fine-tuned to various V+L tasks and achieves state-of-the-art results. The main difference between these models lies in pre-training strategies and cross-modality architecture. To be specific, UNITER \cite{chen2019uniter} and VisualBert \cite{li2019visualbert} applied a single stream of Transformer to jointly learn image-text embedding. LXMERT \cite{tan2019lxmert} and ViLBERT \cite{lu2019vilbert} used two separated Transformer blocks on image and text input and a third fusion Transformer block for cross-modality. Masked language modeling, masked region modeling, image-text matching, word-region alignment, and other pre-training tasks are experimented in those models.

The joint image-text embedding are less discussed in the biomedical domain due to the shortage of large-scale annotated V+L datasets. TandemNet \cite{zhang2017tandemnet}, which adopts a CNN+LSTM architecture, classifies pathological bladder cancer from images with the semantic clues in diagnostic reports. Lau et al. \cite{lau2018dataset} introduce the first manually constructed VQA dataset in radiology, named VQA-RAD. In their study, the performances of 2 well-known CNN+LSTM architecture (MCB \cite{fukui2016multimodal} and SAN \cite{yang2016stacked}) are compared with human radiologists on the VQA-RAD dataset. For drug label identification tasks, the image-text embedding (DLI-IT) \cite{liu2020dli} derived from an VGG+BiLSTM architecture, greatly outperforms image-based or text-based identification methods. Yan et al. \cite{yan2019zhejiang} won the first place in ImageCLEF 2019 VQA-Med \cite{abacha2019vqa} competition using the "VGG16+BERT+MFB" model. Unlike conventional architectures using RNN for text embedding, they encode the semantic questions using the last and penultimate hidden states of the BERT model.

Chest X-Ray datasets are widely used in V+L researches. The joint image-text embedding can be learned from easily accessible Chest X-ray images and free-text radiology reports. TieNet \cite{wang2018tienet}, an end-to-end CNN+RNN architecture, is developed to learn a blend of distinctive image and text representations from the ChestXray14 \cite{wang2017chestx} dataset and are further used for thoracic findings classification and automatic X-ray report generation. Zhang et al. \cite{zhang2019text} achieve better interpretation for OpenI \cite{demner2012design} Chest X-Ray dataset by using dual attention mechanisms to distill visual features with semantic features. Hsu et al. \cite{hsu2018unsupervised} explore supervised and unsupervised methods to learn joint image-text embedding from MIMIC-CXR \cite{johnson2019mimic} dataset. The joint representations are further examined by a document retrieval task. 

We observe that there is a significant gap between the models for joint embedding generation. Transformer-based pre-trained models are widely used in the general domain while in the biomedical domain, CNN-RNN based architectures are still predominating. Our contributions are three-fold: (1) We compare 4 pre-trained Transformer-based V+L models of learning joint image-text embedding from chest radiographs and associated reports; (2) We conduct extrinsic evaluation to compare these V+L models with a pioneering CNN-RNN model (TieNet); (3) We explore the advantage of joint embedding over text-only embedding, pre-training over train-from-scratch and other model components by a detailed ablation study.

\section{Methodology}
\begin{figure}[]
\caption{Overview of applying V+L models to learn joint image-text embedding from CXR images and associated reports for thoracic findings identification.}
\includegraphics[width = \textwidth]{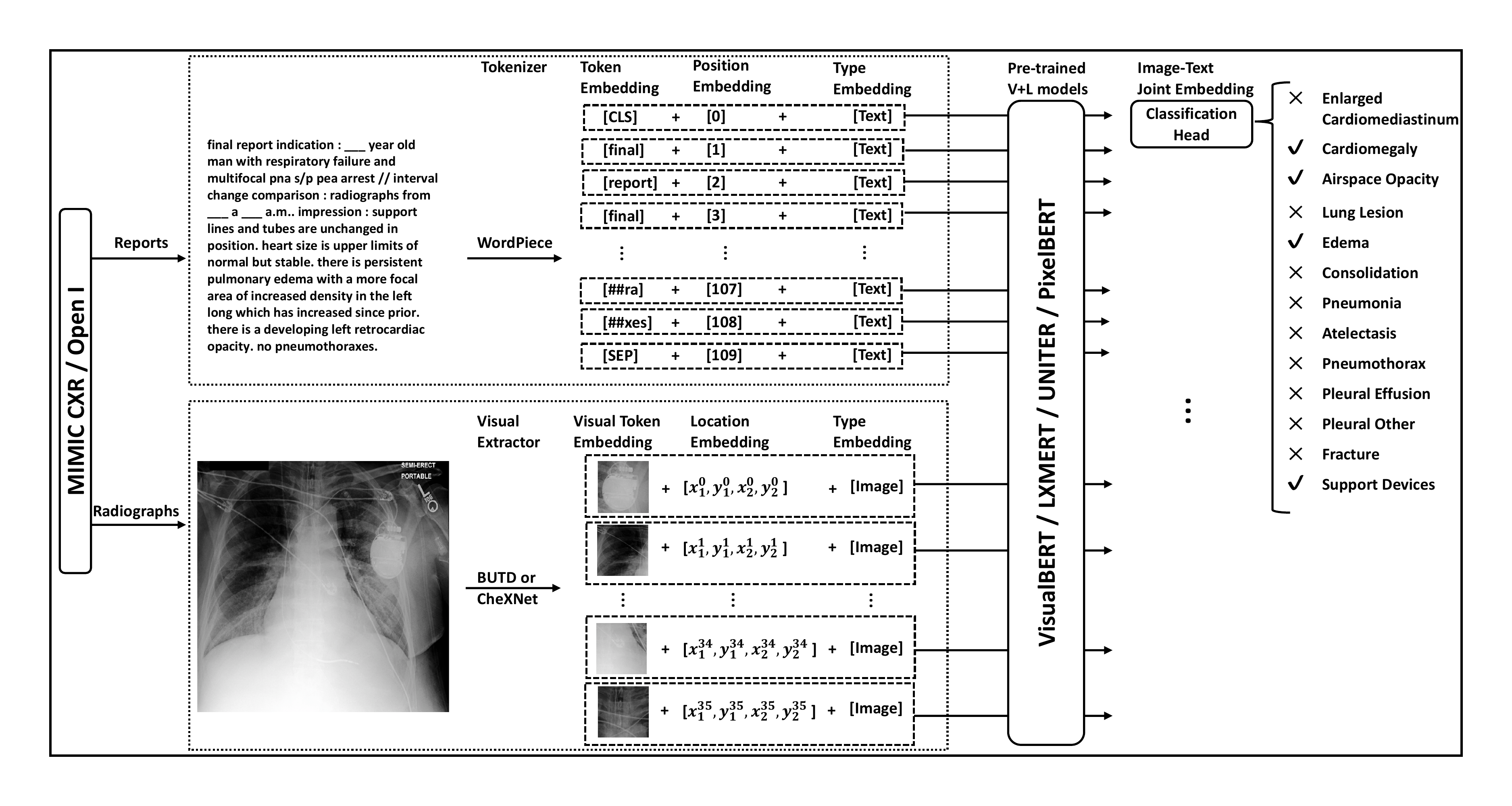}
\label{overview}
\end{figure}

In this section, we introduce how to fine-tune pre-trained V+L models for a multi-label thoracic findings classifier from Chest X-ray images and associated reports. This experiment can be considered as a visual question answering (VQA) task. The workflow is illustrated in Fig. \ref{overview}. 

\subsection{Data}
\subsubsection{MIMIC-CXR}
MIMIC-CXR\cite{johnson2019mimic} is the largest publicly available Chest X-ray dataset that contains 377,110 radiographs and 227,827 free-text radiology associated reports. 13 thoracic findings (detailed in the `Finding' column of Table \ref{mimic}) and `\textit{No Finding}' (absence of all 13 findings) are derived from radiology reports by using 2 auto-annotators: CheXpert \cite{irvin2019chexpert} and NegBio \cite{peng2018negbio}. Each label is categorized into one of four classes: \textit{positive, negative, uncertain and missing}.

We pre-process the data as follows. First, as one radiology report can be associated with multiple radiographs, we only keep the first frontal view radiograph in chronological order. Those reports without any frontal view radiographs are discarded. Then, we re-categorize the labels of radiology reports from 4 classes: \textit{positive, negative, uncertain and missing} to 2 classes: \textit{positive and all others}. This step helps us keep our processed labels consistent with other public chest X-ray datasets (e.g. ChestXRay14, Open I). In total, we have 222,713 image-report pairs and each pair corresponds to 14 binary labels. The sample numbers and prevalence of each finding are shown in the \textit{\#,\%} columns in Table \ref{mimic}. 

\subsection{Open I}
OpenI\cite{demner2016preparing} is another publicly available chest X-ray dataset collected by Indiana University. The dataset contains 3,996 radiology reports associated with 8,121 images. Unlike MIMIC-CXR, which is annotated by auto-annotators, OpenI dataset is labeled by medical professionals manually using Medical Subject Heading \cite{lipscomb2000medical} (MeSH) indexing. We apply the same pre-processing steps as TieNet and yielded 3,684 image-report pairs. Each pair is assigned multiple MeSH terms by human-annotators. For comparison purposes, we only keep the findings that also present in MIMIC-CXR. The positive pair sample numbers and prevalence of selected findings are shown in the \textit{\#,\%} columns in Table \ref{openI}.

\begin{table}[t!]
\caption{A comparison across four pre-trained vision-and-language models and two language representation models.}
\resizebox{\textwidth}{!}{%
\begin{tabular}{@{}llllllll@{}}
\toprule
Model & \begin{tabular}[c]{@{}l@{}}Pretraining\\ Dataset\end{tabular} & Pretraining Tasks* & Visual Encoder & Location Features & Model Architecture** &\begin{tabular}[c]{@{}l@{}}Transformer\\ Streams\end{tabular}  & \begin{tabular}[c]{@{}l@{}}Baseline\\ Performance***\end{tabular} \\ \midrule
\textbf{UNITER} \cite{chen2019uniter}& \begin{tabular}[c]{@{}l@{}}COCO \cite{lin2014microsoft}+VG \cite{krishna2017visual}+\\ CC \cite{sharma2018conceptual}+SBU \cite{zhu2016visual7w}\end{tabular} & \begin{tabular}[c]{@{}l@{}}MLM+MRM+\\ ITM+WRA\end{tabular} & BUTD \cite{anderson2018bottom}& \begin{tabular}[c]{@{}l@{}}{[}top, left, bottom, \\ right, width, height, \\ area{]} of ROIs\end{tabular} & 12 BertLayers & Single & 72.70 (COCO) \\
\textbf{LXMERT}  \cite{tan2019lxmert} & \begin{tabular}[c]{@{}l@{}}COCO+VG+VQA2.0 \cite{antol2015vqa}+\\GQA \cite{hudson2019gqa}+VG-QA \cite{ordonez2011im2text}\end{tabular} & \begin{tabular}[c]{@{}l@{}}MLM+MRM+\\ ITM+IQA\end{tabular} & BUTD & \begin{tabular}[c]{@{}l@{}}{[}top, left, bottom, \\ right{]} of ROIs\end{tabular} & \begin{tabular}[c]{@{}l@{}}9 BertLayers for language,\\ 5 BertLayers for vision,\\ 5 CrossAttLayers for \\ cross-modality\end{tabular} & Double & 72.40 (COCO) \\
\textbf{VisualBERT} \cite{li2019visualbert} & COCO+VQA2.0 & \begin{tabular}[c]{@{}l@{}}MLM+ITM+\\ Task-specific\\ Pretraining\end{tabular} & BUTD & - & 12 BertLayers & Single & 70.80 (COCO) \\
\textbf{PixelBERT} \cite{huang2020pixelbert} & COCO+VG & MLM+ITM & \begin{tabular}[c]{@{}l@{}}ResNeXt \cite{DBLP:journals/corr/XieGDTH16}\\ (Trainable)\end{tabular} & - & 12 BertLayers & Single & 74.45 (COCO) \\
\textbf{BERT} \cite{devlin2018bert} & \begin{tabular}[c]{@{}l@{}}BookCorpus+\\ English-Wiki\\\cite{zhu2015aligning} \end{tabular} & MLM+NSP & - & - & 12 BertLayers & Single & 77.6 (MedNLI) \\
\textbf{ClinicalBERT}  \cite{alsentzer2019publicly} & MIMIC-III \cite{johnson2016mimic}& MLM+NSP & - & - & 12 BertLayers & Single & 80.8 (MedNLI) \\ \bottomrule 
\\
\multicolumn{8}{l}{* \textit{MLM: Mask Languae Modeling; MRM: Mask Region Modeling: ITM: Image-Text Matching; WRA: Word-Region Alignment;}}\\
\multicolumn{8}{l}{\textit{IQA: Image Question Answering; NSP: Next sentence prediction.}}\\
\multicolumn{8}{l}{** \textit{BertLayer is a self-attention layer followed by a feed-forward layer; CrossAttLayer is a cross-attention layer followed by a feed-forward layer.}} \\
\multicolumn{8}{l}{*** \textit{For V+L models, we report the accuracy performed on test-dev of COCO; For language representation  models, we report the accuray performed on MedNLI \cite{romanov2018lessons}}}

\end{tabular}
}
\label{model}
\end{table}

\subsection{Models}
We investigate 4 trending vision-and-language models: VisualBERT, UNITER, LXMERT, and PixelBERT. A detailed comparison of these models is outlined in Table \ref{model}. In summary, all 4 V+L models use the COCO \cite{lin2014microsoft} dataset as their pre-training material. Some other data sources, such as Visual Genome \cite{krishna2017visual}, VQA 2.0 \cite{abacha2019vqa} are added to enrich the training materials. Inspired by the masked language modeling (MLM) and next-sentence prediction of the BERT model, MLM conditioned on images and image-text matching (IMT) are adopted by all 4 V+L models. Only VisualBERT requires task-specific training to improve the performance on downstream tasks. Bottom-up top-down (BUTD) \cite{anderson2018bottom}, adopted by VisualBERT, UNITER, and LXMERT, is the most common region-based visual encoder in V+L models. However, region-based visual encoders are designed for specific visual tasks and are limited by the given categories of the object detection tasks. PixelBERT creatively solves the limitation of region-based visual representation by proposing a CNN-based visual encoder. Note that the self-attention mechanism in the Transformer model is order-less, thus UNITER and LXMERT employ one layer of location features in addition to the original visual features. LXMERT is the only model that has two transformer streams that learns visual and text embedding separately, and further incorporates them with cross-attention layers to achieve joint embedding. 

Preliminary results show that TieNet-R, an attention-based LSTM model that only needs contextual input, also yielded a competitive result. This is in line with expectations since radiology reports are generated by experienced radiologists that contain richer and more interpretative information compared with images. Moreover, the finding labels are auto-annotated by CheXpert, a rule-based text classifier, on radiology reports. Therefore, it is natural that applying text-only embedding methods also achieves a good performance. To this end, BERT and ClinicalBERT \cite{alsentzer2019publicly}, two pre-trained language representation models, are also involved in our study for the purpose of exploring the advantages of image-text joint embedding over text-only embedding. These two models have the same Transformer-based architecture as PixelBERT, VisualBERT, and UNITER. BERT obtained state-of-the-art results on 11 NLP tasks when released. ClinicalBERT, trained on MIMIC-III \cite{johnson2016mimic} dataset, enhances adaptability of BERT for the clinical domain and yields performance improvements on clinical NLP tasks.

\subsection{Implementations}
\begin{table*}[]
\caption{AUCs of identifying 7 thoracic findings from OpenI dataset using TieNet, 4 pre-trained vision-and-language models, and 2 language representation models.}
\centering
\resizebox{0.9\textwidth}{!}{%
\begin{tabular}{@{}lllllllllllll@{}}
\toprule
Findings & \multicolumn{5}{c}{Image+Report} & \multicolumn{1}{c}{} & \multicolumn{3}{c}{Report} &  & \multirow{2}{*}{\#} & \multirow{2}{*}{\%} \\ \cmidrule(lr){2-6} \cmidrule(lr){8-10}
 & \multicolumn{1}{c}{\textbf{TieNet-I+R}} & \multicolumn{1}{c}{\textbf{VisualBERT}} & \multicolumn{1}{c}{\textbf{LXMERT}} & \multicolumn{1}{c}{\textbf{UNITER}} & \multicolumn{1}{c}{\textbf{PixelBERT}} & \multicolumn{1}{c}{\textbf{}} & \multicolumn{1}{c}{\textbf{TieNet-R}} & \multicolumn{1}{c}{\textbf{BERT}} & \multicolumn{1}{c}{\textbf{ClinicalBERT}} &  &  &  \\ \midrule
Cardiomegaly & 0.962 & 0.977 & \textbf{0.980} & 0.978 & 0.970 &  & 0.944 & 0.976 & 0.969 &  & 315 & 8.55 \\
Edema & \textbf{0.995} & 0.982 & \textbf{0.995} & 0.989 & \textbf{0.995} &  & 0.984 & 0.987 & 0.976 &  & 40 & 1.09 \\
Consolidation & 0.989 & 0.996 & 0.993 & \textbf{0.998} & 0.975 &  & 0.969 & 0.978 & 0.982 &  & 28 & 0.76 \\
Pneumonia & \textbf{0.994} & 0.990 & 0.990 & 0.988 & 0.969 &  & 0.983 & 0.970 & 0.982 &  & 36 & 0.98 \\
Atelectasis & 0.972 & \textbf{0.992} & 0.989 & 0.982 & 0.959 &  & 0.981 & 0.927 & 0.947 &  & 293 & 7.95 \\
Pneumothorax & 0.960 & \textbf{0.988} & 0.958 & 0.983 & 0.982 &  & 0.960 & 0.930 & 0.973 &  & 22 & 0.60 \\
Pleural Effusion & 0.977 & \textbf{0.985} & 0.983 & 0.983 & 0.970 &  & 0.968 & 0.961 & 0.976 &  & 140 & 3.80 \\ \midrule
Average & 0.978 & \textbf{0.987} & 0.984 & 0.986 & 0.974 &  & 0.970 & 0.961 & 0.972 &  &  &  \\ \midrule
\#w Average* & 0.971 & \textbf{0.985} & 0.984 & 0.982 & 0.968 &  & 0.965 & 0.956 & 0.964 &  &  &  \\ \bottomrule\\
\multicolumn{13}{l}{* \textit{\#w Averaged denotes the sample number weighted average of AUC.}}
\end{tabular}
}
\label{openI}
\end{table*}

\begin{table*}[!b]
\caption{AUCs of identifying 13 thoracic findings from MIMIC-CXR dataset by fine-tuning 4 pre-trained vision-and-language models and 2 language representation models. Multiple configurations of each model are applied for ablation studies usage.}

\resizebox{\textwidth}{!}{%
\begin{tabular}{@{}llllllllllllllllllll@{}}
\toprule
Findings & \multicolumn{14}{c}{Image+Report} & \multicolumn{1}{c}{} & \multicolumn{2}{c}{Report} & \multicolumn{1}{c}{\multirow{3}{*}{\#}} & \multicolumn{1}{c}{\multirow{3}{*}{\%}} \\ \cmidrule(lr){2-15} \cmidrule(lr){17-18}
 & \multicolumn{3}{c}{\textbf{LXMERT}} & \multicolumn{1}{c}{\textbf{}} & \multicolumn{3}{c}{\textbf{VisualBERT}} & \multicolumn{1}{c}{\textbf{}} & \multicolumn{3}{c}{\textbf{UNITER}} & \multicolumn{1}{c}{\textbf{}} & \multicolumn{2}{c}{\textbf{PixelBERT}} & \textbf{} & \textbf{BERT} & \textbf{\begin{tabular}[c]{@{}l@{}}Clinical-\\ BERT\end{tabular}} & \multicolumn{1}{c}{} & \multicolumn{1}{c}{} \\ \cmidrule(lr){2-4} \cmidrule(lr){6-8} \cmidrule(lr){10-12} \cmidrule(lr){14-15}
\multicolumn{1}{r}{\textit{Config*}} & Default & FS & CB &  & Default & FS & CB &  & Default & FS & CB &  & Default & NT &  &  &  & \multicolumn{1}{c}{} & \multicolumn{1}{c}{} \\ \midrule
\begin{tabular}[c]{@{}l@{}}Enlarged-\\ Cardiomediastinum\end{tabular} & 0.974 & 0.971 & 0.971  &  & \textbf{0.981} & 0.856 & 0.977 &  & 0.979 & 0.866 & 0.962 &  & 0.870 & 0.724 &  & 0.747 & 0.966 & 7,025 & 3.15 \\
Cardiomegaly & 0.987 & 0.988 & 0.988  &  & \textbf{0.991} & 0.968 & 0.990 &  & 0.989 & 0.974 & 0.981 &  & 0.965 & 0.814 &  & 0.954 & 0.979 & 43,931 & 19.73 \\
Airspace Opacity & 0.988 & 0.988 & 0.988 &  & \textbf{0.991} & 0.977 & 0.989 &  & 0.989 & 0.981 & 0.984 &  & 0.971 & 0.922 &  & 0.972 & 0.978 & 50,238 & 22.56 \\
Lung Lesion & 0.983 & 0.963 & 0.963  &  & \textbf{0.985} & 0.947 & 0.983 &  & 0.981 & 0.948 & 0.962 &  & 0.951  & 0.723 &  & 0.911 & 0.972 & 6,312 & 2.83 \\
Edema & 0.988 & 0.989 & 0.988 &  & \textbf{0.991} & 0.975 & 0.990 &  & 0.990 & 0.978 & 0.985 &  & 0.958 & 0.790 &  & 0.956 & 0.979 & 26,309 & 11.81 \\
Consolidation & 0.985 & 0.986 & 0.986 &  & \textbf{0.989} & 0.967 & 0.986 &  & 0.988 & 0.974 & 0.981 &  & 0.973 & 0.772 &  & 0.947 & 0.979 & 10,507 & 4.72 \\
Pneumonia & 0.969 & 0.971 & 0.971 &  & \textbf{0.977} & 0.932 & 0.973 &  & 0.974 & 0.948 & 0.961 &  & 0.942 & 0.741 &  & 0.886 & 0.962 & 16,539 & 7.43 \\
Atelectasis & 0.985 & 0.986 & 0.986 &  & \textbf{0.988} & 0.975 & 0.986 &  & 0.987 & 0.978 & 0.983 &  & 0.971 & 0.797 &  & 0.965 & 0.976 & 44,887 & 20.15 \\
Pneumothorax & 0.990 & 0.985 & 0.985 &  & \textbf{0.992} & 0.975 & 0.991 &  & 0.991 & 0.981 & 0.985 &  & 0.942 & 0.827 &  & 0.942 & 0.979 & 10,432 & 4.68 \\
Pleural Effusion & 0.990 & 0.991 & 0.991 &  & \textbf{0.993} & 0.982 & 0.992 &  & 0.992 & 0.984 & 0.988 &  & 0.972 & 0.945 &  & 0.967 & 0.981 & 52,890 & 23.75 \\
Pleural Others & \textbf{0.983} & 0.969 & 0.969 &  & 0.981 & 0.946 & 0.973 &  & 0.973 & 0.956 & 0.963 &  & 0.930 & 0.735 &  & 0.815 & 0.964 & 1,949 & 0.88 \\
Fracture & \textbf{0.979} & 0.976 & 0.976 &  & 0.976 & 0.960 & 0.975 &  & 0.977 & 0.956 & 0.961 &  & 0.941 & 0.602 &  & 0.873 & 0.958 & 4,425 & 1.99 \\
Support Devices & 0.993 & 0.992 & 0.992 &  & \textbf{0.995} & 0.985 & 0.993 &  & 0.994 & 0.986 & 0.990 &  & 0.979 & 0.949 &  & 0.970 & 0.983 & 65,235 & 29.29 \\ \midrule
Average & 0.984 & 0.981 & 0.981 & \textbf{} & \textbf{0.987} & 0.957 & 0.984 & \textbf{} & 0.985 & 0.962 & 0.976 & \textbf{} & 0.953 & 0.795 & \textbf{} & 0.916 & 0.974 &  &  \\ \bottomrule
\\
\multicolumn{20}{l}{* \textit{Configurations for ablation studies: 
Default: default configuration; 
FS: training from scratch without loading pre-trained model weights;}}\\
\multicolumn{20}{l}{\textit{CB: replacing the tokenizer, vocabulary and embedding layer with those of ClinicalBERT; 
NT: freezing weights of the visual backbone.}}
\end{tabular}
}
\label{mimic}
\end{table*}

We first fine-tune the 4 V+L models to learn a joint image-text embedding (the last hidden-state of [CLS] token) from the training set of MIMIC-CXR. A classification head is added on top of the joint embedding to generate the probability estimates of 13 thoracic findings. We use the held-out testing set of MIMIC-CXR for internal evaluation and entire OpenI dataset for external evaluation. It should be noted that we do not follow the official splits of MIMIC-CXR. Instead, we perform stratified split and use 80\% of the entire dataset as the training set, 10\% as developing set for hyper-parameter selection, and the rest 10\% for held-out internal evaluation. TieNet \cite{wang2018tienet}, a pioneering joint embedding model used RNN-CNN architecture, is set as a baseline in our external evaluation. Areas Under Receiver Operating Curves (AUC) is computed based on the probability estimates and annotated labels to evaluate the model performance. We follow the evaluations in TieNet and report averaged AUC and sample number weighted average of AUC for comparison.

For detailed implementation, each text sequence is truncated or padded to 128 tokens in length. We apply a special edition of BUTD visual encoder for UNITER, LXMERT and VisualBERT, which consistently extracts 36 objects from a single image. As the pre-trained weights of PixelBERT are not open-sourced, we adopt CheXNet \cite{DBLP:journals/corr/abs-1711-05225} as a substitute of the ResNeXt as the visual backbone and initialize the BertLayers with pre-trained BERT weights. The feature map (7x7x1024) of CheXNet is first flattened by spatial dimensions (49x1024) and then down-sampled to 36 1024-long visual features. We follow the same setting as \cite{huang2020pixelbert} that trains the PixelBERT for 18 epochs. All other 3 V+L models as well as BERT and ClinicalBERT are fine-tuned for 6 epochs as the pre-trained weights can be loaded. SGD optimizer, with weight decay 5e-4 and learning rate 0.01 scheduled decay by 10 at the 12\textsuperscript{th} and 16\textsuperscript{th} epoch, is used to optimize the CheXNet backbone in PixelBERT. AdamW is used to optimize the Transformer block(s) in all models. Each model can be fit into 1 Tesla K40 GPU when using a batch size of 16. Our implementation can be found at: \textit{https://github.com/YIKUAN8/Transformers-VQA}

\section{Results} 

\begin{figure}[]
\caption{Visualization of the selected attention maps learned by the VisualBERT model.}
\includegraphics[width = \textwidth]
{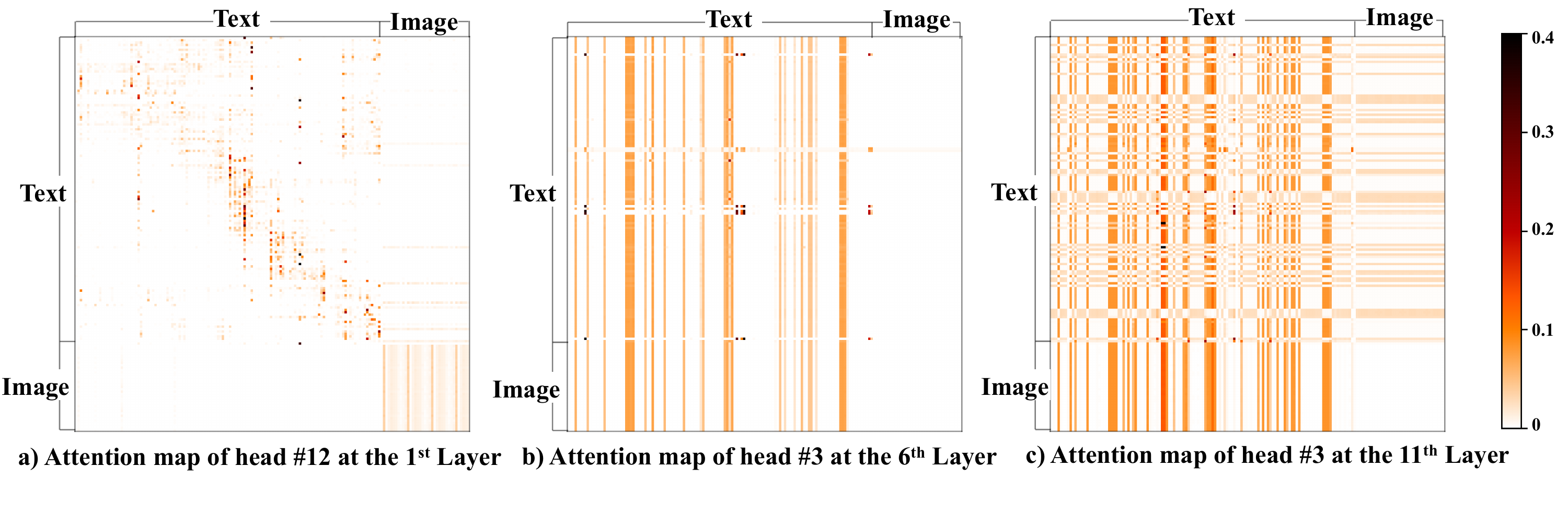}
\label{vis}
\end{figure}

The model performance on OpenI dataset are shown in Table \ref{openI}. All V+L models yield a high AUC of over 0.958 for all findings. VisualBERT, LXMERT and UNITER achieve higher averaged AUC (0.984+) and sample number weighted AUC (0.982+) than TieNet-I+R (0.978 and 0.971) in OpenI dataset. While PixelBERT, without any pre-training, performs slightly worse than TieNet-I+R. The supreme performance demonstrates all V+L models' great adaptability to biomedical domain. When digging into a particular finding, TieNet only outperforms V+L models in the classification of \textit{edema} and \textit{pneumonia}. 

We use the held-out test split of MIMIC-CXR to make a more comprehensive comparison of 4 V+L models. The results are revealed in the \textit{default} column of Table \ref{mimic}. VisualBERT yields the best averaged AUC of 0.987, followed by 0.985 using UNITER and 0.984 using LXMERT. PixelBERT only achieves a moderate AUC of 0.953 due to the absence of the pre-training. When comparing the performance on different findings, we surprisingly find that VisualBERT achieves the best results on 11 out of 13 findings. Although VisualBERT adopts the least pre-training tasks and the simplest model architecture, we believe the introduction of task-specific training for VQA contributes to this optimal performance. LXMERT outperforms others in \textit{Pleural Others} and \textit{Fracture}, which are the two least prevalent findings. We also notice that it is harder to identify \textit{Enlarged Cardiomediastinum} compared with other abnormalities. We attribute this to the absence of lateral view X-ray images in our training set in that the cardiomediastinum contour is easier to recognize from a lateral view. Moreover, it is as expected that UNITER does not have a supreme performance in our study as we adopt the base version of UNITER. UNITER\textsubscript{larger} is much larger than UNITER\textsubscript{base} in both number of layers and trainable parameters. We believe that UNITER\textsubscript{larger} can achieve better results given its phenomenal performance in general domain. 

Next, we compare the advantage of using image-text joint embedding over text-only embedding. When using joint embedding, TieNet only receives less than 1\% performance improvement over the text-only embedding in the OpenI dataset. While, VisualBERT, LXMERT, and UNITER receives more than 2 \%  improvement over BERT and 1 \% over ClinicalBERT, respectively. This larger performance improvement demonstrates that transformer-based models are better at image-text interaction compared with the CNN-RNN architecture. In the test split of MIMIC-CXR, the improvement of V+L models over BERT model is even larger (more than 6\%).

Additionally, although VisualBERT, LXMERT, and UNITER have different model architectures, embedding, and pre-training strategies, they all yield comparable results in both internal held-out evaluation and external data resources. This is in line with expectation because these V+L models all have supreme performance and rank top 10 in the general domain (evaluated by coco test-dev). This also indicates that although BUTD is pre-trained in general domain from Visual Genome, it could also extract useful ROIs from biomedical images.

\section{Discussion}
\begin{figure}[]
\caption{Results of three classification samples. Top 10 tokens with highest attention scores are highlighted in each sample. Words in grey are truncated during tokenization.}
\centering
\includegraphics[width = 0.5\textwidth]{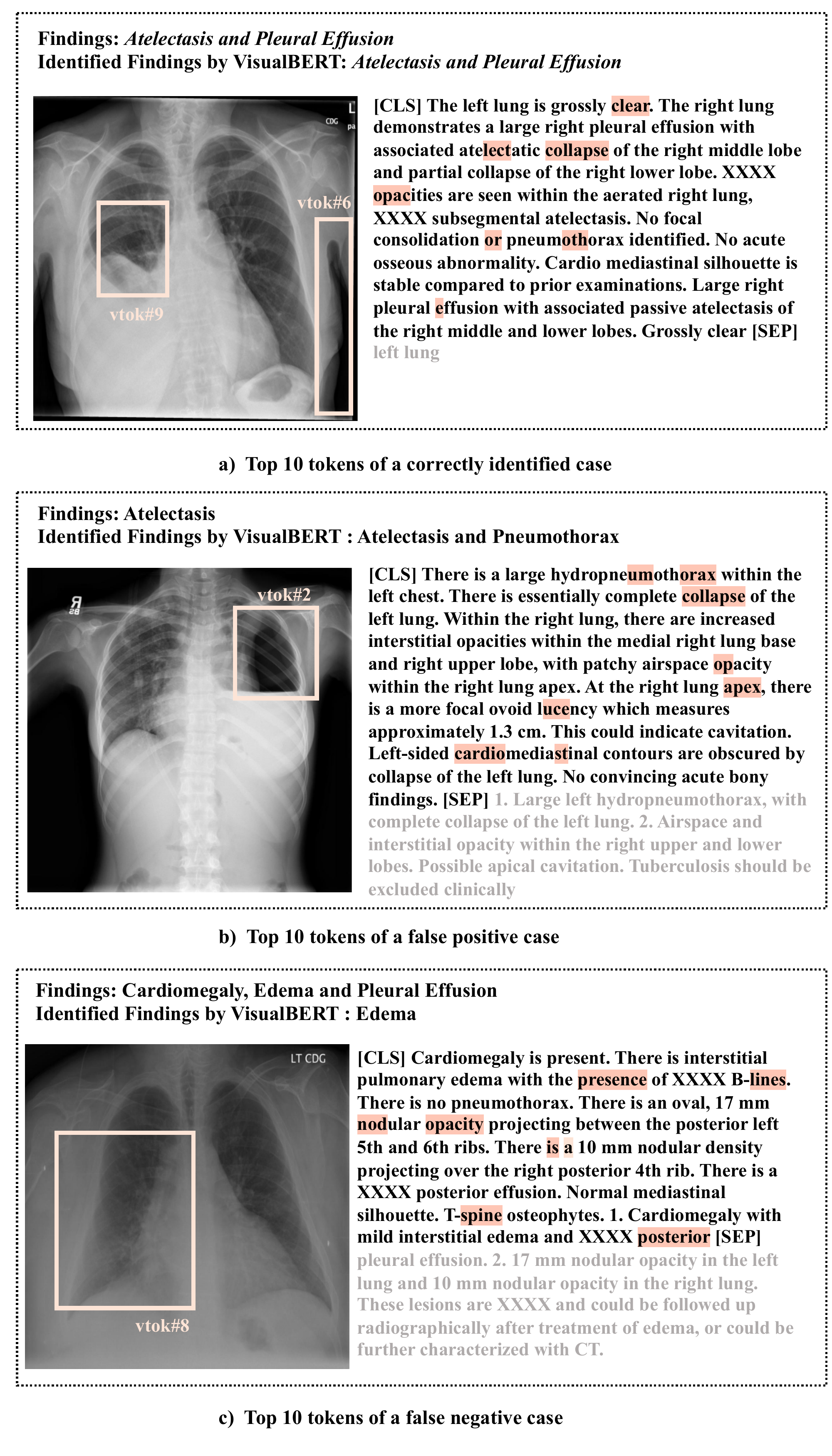}
\label{error}
\end{figure}

In this section, we first present ablation studies to analyze the contribution of certain model components. Then, we use visualization to illustrate the attention details. Finally, we discuss the limitations of our current work and our plans for further studies.
\begin{table*}[]
\caption{Layer-level modality importance scores. A lower score indicates that [CLS] token spends less attention on the modality.}
\centering
\resizebox{0.75\textwidth}{!}{%
\begin{tabular}{@{}lllllllllllll@{}}
\toprule
\textbf{Layer\#} & \multicolumn{1}{c}{\textbf{1}} & \multicolumn{1}{c}{\textbf{2}} & \multicolumn{1}{c}{\textbf{3}} & \multicolumn{1}{c}{\textbf{4}} & \multicolumn{1}{c}{\textbf{5}} & \multicolumn{1}{c}{\textbf{6}} & \multicolumn{1}{c}{\textbf{7}} & \multicolumn{1}{c}{\textbf{8}} & \multicolumn{1}{c}{\textbf{9}} & \multicolumn{1}{c}{\textbf{10}} & \multicolumn{1}{c}{\textbf{11}} & \multicolumn{1}{c}{\textbf{12}} \\ \midrule
Textual Modality Importance & 0.828 & 0.993 & 0.941 & 0.824 & 0.844 & 0.908 & 0.724 & 0.733 & 0.812 & 0.864 & 0.922 & 0.673 \\
Visual Modality Importance & 0.170 & 0.007 & 0.055 & 0.103 & 0.147 & 0.082 & 0.254 & 0.206 & 0.172 & 0.134 & 0.075 & 0.324 \\ \bottomrule
\end{tabular}%
}
\label{mi}
\end{table*}

\subsection{Pre-training versus Training From Scratch}
We explore whether pre-training strategies and tasks that are utilized in the general domain contribute to our joint representation learning in the biomedical domain. As a comparative study, we do not load the pre-trained weights and randomly initialize the weights for VisualBERT, LXMERT, and UNITER. These 3 V+L models are directly trained to classify thoracic findings without any pre-training. We train these models for 12 epochs. The results of training from scratch are shown in the \textit{FS} columns of Table \ref{openI} and are compared to the \textit{Default} columns. The pre-training procedure improves the averaged AUC of VisualBERT by 3\%, UNITER by 2.3\% and LXMERT by 2\%. This confirms that the pre-training tasks contribute to generating more meaningful joint embedding. In the contrast, PixelBERT achieves state-of-the-art results in many NLP tasks in general domain, our train-from-scratch PixelBERT underperforms the other 3 pre-trained model significantly. This also indicates that pre-training is of great importance to multimodal representation learning.

\subsection{BERT versus ClinicalBERT}
We observe that ClinicalBERT outperforms BERT by 5.8\% in MIMIC-CXR and 1.1\% in OpenI. It is as expected in that ClinicalBERT is pre-trained using MIMIC-III which is also released by PhysioNet. Compared with BERT, ClinicalBERT adapts better to biomedical data. Inspired by this, we replace the BERT vocabulary and embedding layer in VisualBERT, LXMERT and UNITER by clinicalBERT. In terms of implementation, we load all the pre-trained parameters except the text embedding layer to those 3 V+L models. We also train the models for 12 epochs in hope of more training epochs will help V+L models adapt to the new vocabulary and embedding layer. The results of using ClinicalBERT components are shown in the \textit{CB} columns of Table \ref{mimic}. Compared with the default configuration, we do not observe any performance boosting. We believe the 12 training epochs are still not sufficient to adjust the parameters to adapt the a new language representation model.

\subsection{Trainable Visual Backbone versus Frozen Visual Backbone}
We are also interested in whether a trainable visual CNN backbone is beneficial. This experiment is also conducted in the original work of PixelBERT. We adopt another configuration of PixelBERT by freezing all the layers of CheXNet, the visual backbone CNN. All other hyper-parameters and technical details remain the same. The results of the PixelBERT with a frozen CNN backbone can be found in the \textit{NT} column of Table \ref{mimic}. PixelBERT with frozen CheXNet only yields a moderate averaged AUC of less than 0.8, which underperforms the default configuration of more than 15\%. This indicates that it is very important to adjust the weights of the visual backbone model when applying feature map of the CNN model to generate image representations.

\subsection{Visualization}
Similar to \cite{kovaleva2019revealing}, we first visualize selected attention maps of the best performed VisualBERT to illustrate the attention mechanism. In a), b) and c) of Fig. \ref{vis}, we observe several interesting patterns. Fig. \ref{vis}a shows a block pattern. Only intra-modality (text self-attention and image self-attention) can be observed in this attention map. Fig. \ref{vis}b has a vertical pattern that illustrates Text-to-Image and Text-to-Text attentions in this layer. Besides, we can see some special tokens are heavily attended. Fig. \ref{vis}c visualizes the attention scores of a head in the penultimate layer. Text-to-Image, Image-to-Text and Text self-attention can be found in this attention map. When comparing the difference of these 3 attention maps, we observe there is a trending that the deeper layers demonstrate more interaction between image and text, while self-attention predominate in the beginning several layers. Furthermore, to quantitatively analyze the [CLS] attention trace, we calculate the layer-level \textit{Modality Importance (MI) scores} introduced in \cite{cao2020behind} of textual and visual modality, respectively. The result is shown in Table \ref{mi}. We can see that layer-level MI scores for text modality are significantly higher than those for visual modality, especially at intermediate layers. This suggests that more attention heads learn useful knowledge from the textual modality than image modality. 

% \usepackage{booktabs}
% \usepackage{graphicx}

\subsection{Error Analysis}
For error analysis analysis, the top 10 tokens across all heads and layers with highest attention scores attended by the [CLS] token are highlighted in the Fig. \ref{error}. All three cases are sampled from OpenI. In Fig. \ref{error}a, a sample diagnosed with \textit{Atelectasis, Pleural Effusion} is correctly classified. We can observe that keywords, such as \textit{atelectatic collapse, opacity, effusion}, display high attention by the classification token. Visual token \#9, which precisely targets the contour of the collapse of the right middle lobe, also greatly contributes to the classification. On the contrary, Fig. \ref{error}b illustrates a false negative sample. This sample is wrongly classified to positive \textit{Pneumothorax} by our model. As shown in the radiology report of this sample, there is a large \textit{hydropneumothorax} within the left lung. Given that all V+L models use WordPiece, which is a subword algorithm, \textit{hydropneumothorax} is tokenized to a superset of \textit{pneumothorax}. The tokenization mechanism of V+L models results in this kind of false positive classification. Fig. \ref{error}c demonstrates a false negative sample. This patient is diagnosed with \textit{Cardiomegaly, Edema, and Pleural Effusion}. However, only \textit{Edema} is identified by our classifier. When digging into the specific tokens, no finding-related words but \textit{nodular opacity} receive a high attention score. \textit{Pleural Effusion} in the context is truncated as we set the maximum text sequence to 128. While, we can see that visual token \#8 precisely targets thoracic-spine and \textit{T-spine} in radiology report also has a high attention score. This demonstrates the strong image-text alignment ability of V+L models.
\subsection{Limitation and Future Work}
Our study has several limitations, each of which brings a great opportunity for future work. First, we barely tune the hyper-parameters or CNN backbone architectures of the V+L models, which has the potential of achieving better performance. For example, we could apply deeper CNN backbones, such as ResNet152 to extract visual features for PixelBERT. Additionally, due to the limitation of computation resources, we are unable to implement domain-specific pre-training. According to the previous studies \cite{li2019visualbert, tan2019lxmert}, domain-specific pre-training will enhance the model's adaptability and therefore improve the performance of domain-specific tasks. Finally, we only apply the pre-trained V+L models to chest X-ray datasets. We are planning to examine these models on a wider range of biomedical V+L datasets and tasks, such as BCIDR \cite{zhang2017tandemnet}, VQA-Med \cite{abacha2019vqa} with more configurations as our next step.

\section{Conclusion}
In this study, we examine the ability of four transformer-based V+L models in learning joint image-text embedding from chest X-ray images and associated reports. The extrinsic evaluation shows that pre-trained V+L models outperform traditional CNN-RNN methods. Ablation studies demonstrate the advantage of joint embedding over text embedding and the benefits of the pre-training procedures. Future studies will focus on pre-training V+L models using biomedical dataset and down-stream the pre-trained models to more biomedical V+L tasks.

\section{Acknowledgment}
This study is supported in part by NIH grant 1R01LM013337.

\bibliographystyle{unsrt}  
\bibliography{references}  %%% Remove comment to use the external .bib file (using bibtex).


\end{document}